\documentstyle[aclap]{article}

\input{psfig}
\title{\vspace{-0.5in}Statistical Decision-Tree Models for
Parsing\thanks{This work was sponsored by the Advanced Research
Projects Agency, contract DABT63-94-C-0062.  It does not reflect the
position or the policy of the U.S.\ Government, and no official
endorsement should be inferred.  Thanks to the members of the IBM
Speech Recognition Group for their significant contributions to this
work.}}

\author{David M. Magerman\\
Bolt Beranek and Newman Inc.\\
70 Fawcett Street, Room 15/148\\
Cambridge, MA 02138, USA\\
{\tt magerman@bbn.com}}

\begin{document}
\maketitle
\vspace{-0.5in}
\begin{abstract}

Syntactic natural language parsers have shown themselves to be
inadequate for processing highly-ambiguous large-vocabulary text, as
is evidenced by their poor performance on domains like the Wall Street
Journal, and by the movement away from parsing-based approaches to
text-processing in general.  In this paper, I describe SPATTER, a
statistical parser based on decision-tree learning techniques which
constructs a complete parse for every sentence and achieves accuracy
rates far better than any published result.  This work is based on the
following premises: (1) grammars are too complex and detailed to
develop manually for most interesting domains; (2) parsing models must
rely heavily on lexical and contextual information to analyze
sentences accurately; and (3) existing {$n$}-gram modeling techniques
are inadequate for parsing models.  In experiments comparing SPATTER
with IBM's computer manuals parser, SPATTER significantly outperforms
the grammar-based parser.  Evaluating SPATTER against the Penn
Treebank Wall Street Journal corpus using the PARSEVAL measures,
SPATTER achieves 86\% precision, 86\% recall, and 1.3 crossing
brackets per sentence for sentences of 40 words or less, and 91\%
precision, 90\% recall, and 0.5 crossing brackets for sentences
between 10 and 20 words in length.

\end{abstract}

\section{Introduction}

Parsing a natural language sentence can be viewed as making a sequence
of disambiguation decisions: determining the part-of-speech of the
words, choosing between possible constituent structures, and selecting
labels for the constituents.  Traditionally, disambiguation problems
in parsing have been addressed by enumerating possibilities and
explicitly declaring knowledge which might aid the disambiguation
process.  However, these approaches have proved too brittle for most
interesting natural language problems.

This work addresses the problem of automatically discovering the
disambiguation criteria for all of the decisions made during the
parsing process, given the set of possible features which can act as
disambiguators.  The candidate disambiguators are the words in the
sentence, relationships among the words, and relationships among
constituents already constructed in the parsing process.

Since most natural language rules are not absolute, the disambiguation
criteria discovered in this work are never applied deterministically.
Instead, all decisions are pursued non-deterministically according to
the probability of each choice.  These probabilities are estimated
using statistical decision tree models.  The probability of a complete
parse tree ($T$) of a sentence ($S$) is the product of each decision
($d_i$) conditioned on all previous decisions:
\[ P(T|S) = \prod_{d_i\in T}P(d_i|d_{i-1}d_{i-2}\ldots d_1S). \]
Each decision sequence constructs a unique parse, and the parser
selects the parse whose decision sequence yields the highest
cumulative probability.  By combining a stack decoder search with a
breadth-first algorithm with probabilistic pruning, it is possible to
identify the highest-probability parse for any sentence using a
reasonable amount of memory and time.

The claim of this work is that statistics from a large corpus of
parsed sentences combined with information-theoretic classification
and training algorithms can produce an accurate natural language
parser without the aid of a complicated knowledge base or grammar.
This claim is justified by constructing a parser, called SPATTER
(Statistical PATTErn Recognizer), based on very limited linguistic
information, and comparing its performance to a state-of-the-art
grammar-based parser on a common task.  It remains to be shown that an
accurate broad-coverage parser can improve the performance of a text
processing application.  This will be the subject of future
experiments.

One of the important points of this work is that statistical models of
natural language should not be restricted to simple,
context-insensitive models.  In a problem like parsing, where
long-distance lexical information is crucial to disambiguate
interpretations accurately, local models like probabilistic
context-free grammars are inadequate.  This work illustrates that
existing decision-tree technology can be used to construct and
estimate models which selectively choose elements of the context which
contribute to disambiguation decisions, and which have few enough
parameters to be trained using existing resources.

I begin by describing decision-tree modeling, showing that
decision-tree models are equivalent to interpolated {$n$}-gram models.
Then I briefly describe the training and parsing procedures used in
SPATTER.  Finally, I present some results of experiments comparing
SPATTER with a grammarian's rule-based statistical parser, along with
more recent results showing SPATTER applied to the Wall Street Journal
domain.

\section{Decision-Tree Modeling}

Much of the work in this paper depends on replacing human
decision-making skills with automatic decision-making algorithms.  The
decisions under consideration involve identifying constituents and
constituent labels in natural language sentences.  Grammarians, the
human decision-makers in parsing, solve this problem by enumerating
the features of a sentence which affect the disambiguation decisions
and indicating which parse to select based on the feature values.  The
grammarian is accomplishing two critical tasks: identifying the
features which are relevant to each decision, and deciding which
choice to select based on the values of the relevant features.

Decision-tree classification algorithms account for both of these
tasks, and they also accomplish a third task which grammarians
classically find difficult.  By assigning a probability distribution
to the possible choices, decision trees provide a ranking
system which not only specifies the order of preference for the
possible choices, but also gives a measure of the relative likelihood
that each choice is the one which should be selected.

\subsection{What is a Decision Tree?}

A decision tree is a decision-making device which assigns a
probability to each of the possible choices based on the context of
the decision: $P(f|h),$ where $f$ is an element of the {\em future}
vocabulary (the set of choices) and $h$ is a {\em history} (the
context of the decision).  This probability $P(f|h)$ is determined by
asking a sequence of questions $q_1q_2\ldots~q_n$ about the context,
where the $i$th question asked is uniquely determined by the answers
to the $i-1$ previous questions.

\begin{centering}
\begin{figure}[tbhp]
\centerline{\psfig{figure=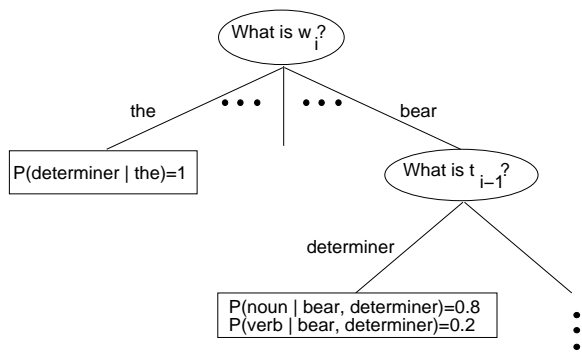,width=3in}}
\caption{Partially-grown decision tree for part-of-speech
tagging.\label{tagtree}}
\end{figure}
\end{centering}

For instance, consider the part-of-speech tagging problem.  The first
question a decision tree might ask is:
\begin{description}
\item[1.] What is the word being tagged?
\end{description}
If the answer is {\em the,} then the decision tree needs to ask no
more questions; it is clear that the decision tree should assign the
tag $f=\mbox{\em determiner}$ with probability 1.  If, instead, the answer
to question 1 is {\em bear,} the decision tree might next ask the
question:
\begin{description}
\item[2.] What is the tag of the previous word?
\end{description}
If the answer to question 2 is {\em determiner,} the decision tree
might stop asking questions and assign the tag $f=\mbox{\em noun}$ with
very high probability, and the tag $f=\mbox{\em verb}$ with much lower
probability.  However, if the answer to question 2 is {\em noun}, the
decision tree would need to ask still more questions to get a good
estimate of the probability of the tagging decision.  The decision
tree described in this paragraph is shown in Figure~\ref{tagtree}.

Each question asked by the decision tree is represented by a {\em tree
node} (an oval in the figure) and the possible answers to this
question are associated with branches emanating from the node.  Each
node defines a probability distribution on the space of possible
decisions.  A node at which the decision tree stops asking questions
is a {\em leaf node.}  The leaf nodes represent the unique states in
the decision-making problem, i.e. all contexts which lead to the same
leaf node have the same probability distribution for the decision.

\subsection{Decision Trees vs.\ {$n$}-grams}

A decision-tree model is not really very different from an
interpolated {$n$}-gram model.  In fact, they are equivalent in
representational power.  The main differences between the two modeling
techniques are how the models are parameterized and how the parameters
are estimated.

\subsubsection{Model Parameterization}

First, let's be very clear on what we mean by an {$n$}-gram model.
Usually, an {$n$}-gram model refers to a Markov process where the
probability of a particular token being generating is dependent on the
values of the previous $n-1$ tokens generated by the same process.  By
this definition, an {$n$}-gram model has $|W|^n$ parameters, where
$|W|$ is the number of unique tokens generated by the process.

However, here let's define an {$n$}-gram model more loosely as a model
which defines a probability distribution on a random variable given
the values of {$n-1$} random variables,
$P(f|h_{1}h_{2}\ldots~h_{n-1}).$ There is no assumption in the
definition that any of the random variables $F$ or $H_i$ range over
the same vocabulary.  The number of parameters in this {$n$}-gram
model is $|F|\prod|H_i|.$

Using this definition, an {$n$}-gram model can be represented by a
decision-tree model with $n-1$ questions.  For instance, the
part-of-speech tagging model $P(t_i|w_{i}t_{i-1}t_{i-2})$ can be
interpreted as a {$4$}-gram model, where $H_1$ is the variable
denoting the word being tagged, $H_2$ is the variable denoting the tag
of the previous word, and $H_3$ is the variable denoting the tag of
the word two words back.  Hence, this {$4$}-gram tagging model is the
same as a decision-tree model which always asks the sequence of 3
questions:
\begin{enumerate}
\item What is the word being tagged?
\item What is the tag of the previous word?
\item What is the tag of the word two words back?
\end{enumerate}

But can a decision-tree model be represented by an {$n$}-gram model?
No, but it can be represented by an {\em interpolated} {$n$}-gram
model.  The proof of this assertion is given in the next section.

\subsubsection{Model Estimation}

The standard approach to estimating an {$n$}-gram model is a two step
process.  The first step is to count the number of occurrences of each
{$n$}-gram from a training corpus.  This process determines the
empirical distribution,
\[ P(f|h_{1}h_{2}\ldots~h_{n-1}) =
\frac{\mbox{Count}(h_{1}h_{2}\ldots~h_{n-1}f)}{\mbox{Count}(h_{1}h_{2}\ldots~h_{n-1})}
\]
The second step is smoothing the empirical distribution using a
separate, held-out corpus .  This step improves the empirical
distribution by finding statistically unreliable parameter estimates
and adjusting them based on more reliable information.

A commonly-used technique for smoothing is deleted interpolation.
Deleted interpolation estimates a model
$\tilde{P}(f|h_{1}h_{2}\ldots~h_{n-1})$ by using a linear combination
of empirical models $P(f|h_{k_1}h_{k_2}\ldots~h_{k_m}),$ where $m~<~n$
and $k_{i-1}~<~k_{i}~<~n$ for all $i\leq~m.$ For example, a model
$\tilde{P}(f|h_1h_2h_3)$ might be interpolated as follows:
\begin{eqnarray*}
\lefteqn{\tilde{P}(f|h_1h_2h_3) =}\\
 & & \lambda_1(h_1h_2h_3)P(f|h_1h_2h_3) +\\
 & & \lambda_2(h_1h_2h_3)P(f|h_1h_2) + \lambda_3(h_1h_2h_3)P(f|h_1h_3)+\\
 & & \lambda_4(h_1h_2h_3)P(f|h_2h_3) + \lambda_5(h_1h_2h_3)P(f|h_1h_2)+ \\
 & & \lambda_6(h_1h_2h_3)P(f|h_1) + \lambda_7(h_1h_2h_3)P(f|h_2) +\\
 & & \lambda_8(h_1h_2h_3)P(f|h_3)
\end{eqnarray*}
where $\sum\lambda_i(h_1h_2h_3)=1$ for all histories $h_1h_2h_3.$ The
optimal values for the $\lambda_i$ functions can be estimated using
the forward-backward algorithm \cite{Baum}.

A decision-tree model can be represented by an interpolated {$n$}-gram
model as follows.  A leaf node in a decision tree can be represented
by the sequence of question answers, or history values, which leads
the decision tree to that leaf.  Thus, a leaf node defines a
probability distribution based on values of those questions:
$P(f|h_{k_1}h_{k_2}\ldots~h_{k_m}),$ where $m~<~n$ and
$k_{i-1}~<~k_{i}~<~n,$ and where $h_{k_i}$ is the answer to one of the
questions asked on the path from the root to the leaf.\footnote{Note
that in a decision tree, the leaf distribution is not affected by the
order in which questions are asked.  Asking about $h_1$ followed by
$h_2$ yields the same future distribution as asking about $h_2$
followed by $h_1.$} But this is the same as one of the terms in the
interpolated {$n$}-gram model.  So, a decision tree can be defined as
an interpolated {$n$}-gram model where the $\lambda_i$ function is
defined as:
\[
\lambda_i(h_{k_1}h_{k_2}\ldots~h_{k_m}) = \left\{
\begin{array}{cl}
1 & \mbox{ if $h_{k_1}h_{k_2}\ldots~h_{k_m}$ is a leaf,}\\
0 & \mbox{ otherwise.}
\end{array} \right.
\]

\subsection{Decision-Tree Algorithms}

The point of showing the equivalence between {$n$}-gram models and
decision-tree models is to make clear that the power of decision-tree
models is not in their expressiveness, but instead in how they can be
automatically acquired for very large modeling problems.  As $n$
grows, the parameter space for an {$n$}-gram model grows
exponentially, and it quickly becomes computationally infeasible to
estimate the smoothed model using deleted interpolation.  Also, as $n$
grows large, the likelihood that the deleted interpolation process
will converge to an optimal or even near-optimal parameter setting
becomes vanishingly small.

On the other hand, the decision-tree learning algorithm increases the
size of a model only as the training data allows.  Thus, it can
consider very large history spaces, i.e. {$n$}-gram models with very
large $n.$ Regardless of the value of $n,$ the number of parameters in
the resulting model will remain relatively constant, depending mostly
on the number of training examples.

The leaf distributions in decision trees are empirical estimates,
i.e. relative-frequency counts from the training data.  Unfortunately,
they assign probability zero to events which can possibly occur.
Therefore, just as it is necessary to smooth empirical {$n$}-gram
models, it is also necessary to smooth empirical decision-tree models.

The decision-tree learning algorithms used in this work were developed
over the past 15 years by the IBM Speech Recognition group
\cite{treelm}.  The growing algorithm is an adaptation of the CART
algorithm in \cite{CART}.  For detailed descriptions and discussions
of the decision-tree algorithms used in this work, see
\cite{mythesis}.

An important point which has been omitted from this discussion of
decision trees is the fact that only binary questions are used in
these decision trees.  A question which has $k$ values is decomposed
into a sequence of binary questions using a classification tree on
those $k$ values.  For example, a question about a word is represented
as 30 binary questions.  These 30 questions are determined by growing
a classification tree on the word vocabulary as described in
\cite{ngrams}.  The 30 questions represent 30 different binary
partitions of the word vocabulary, and these questions are defined
such that it is possible to identify each word by asking all 30
questions.  For more discussion of the use of binary decision-tree
questions, see \cite{mythesis}.

\section{SPATTER Parsing}

The SPATTER parsing algorithm is based on interpreting parsing as a
statistical pattern recognition process.  A parse tree for a sentence
is constructed by starting with the sentence's words as leaves of a
tree structure, and labeling and extending nodes these nodes until a
single-rooted, labeled tree is constructed.  This pattern recognition
process is driven by the decision-tree models described in the
previous section.

\subsection{SPATTER Representation}

A parse tree can be viewed as an {$n$}-ary branching tree, with each
node in a tree labeled by either a non-terminal label or a
part-of-speech label.  If a parse tree is interpreted as a geometric
pattern, a constituent is no more than a set of edges which meet at
the same tree node.  For instance, the noun phrase, ``a brown cow,''
consists of an edge extending to the right from ``a,'' an edge
extending to the left from ``cow,'' and an edge extending straight up
from ``brown''.

\begin{centering}
\begin{figure}[tbhp]
\centerline{\psfig{figure=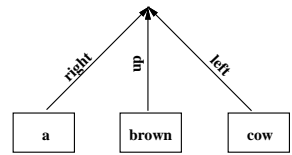,height=0.8in}}
\caption{Representation of constituent and labeling of extensions in
SPATTER.\label{abrowncow}}
\end{figure}
\end{centering}

In SPATTER, a parse tree is encoded in terms of four elementary
components, or {\em features}: words, tags, labels, and extensions.
Each feature has a fixed vocabulary, with each element of a given
feature vocabulary having a unique representation.  The word feature
can take on any value of any word.  The tag feature can take on any
value in the part-of-speech tag set.  The label feature can take on
any value in the non-terminal set.  The extension can take on any of
the following five values:
\begin{description}
\item[right] - the node is the first child of a constituent;
\item[left] - the node is the last child of a constituent;
\item[up] - the node is neither the first nor the last child of a
constituent;
\item[unary] - the node is a child of a unary constituent;
\item[root] - the node is the root of the tree.
\end{description}

For an $n$ word sentence, a parse tree has $n$ leaf nodes, where the
word feature value of the $i$th leaf node is the $i$th word in the
sentence.  The word feature value of the internal nodes is intended to
contain the lexical head of the node's constituent.  A deterministic
lookup table based on the label of the internal node and the labels of
the children is used to approximate this linguistic notion.

\begin{centering}
\begin{figure}[tbhp]
\centerline{\psfig{figure=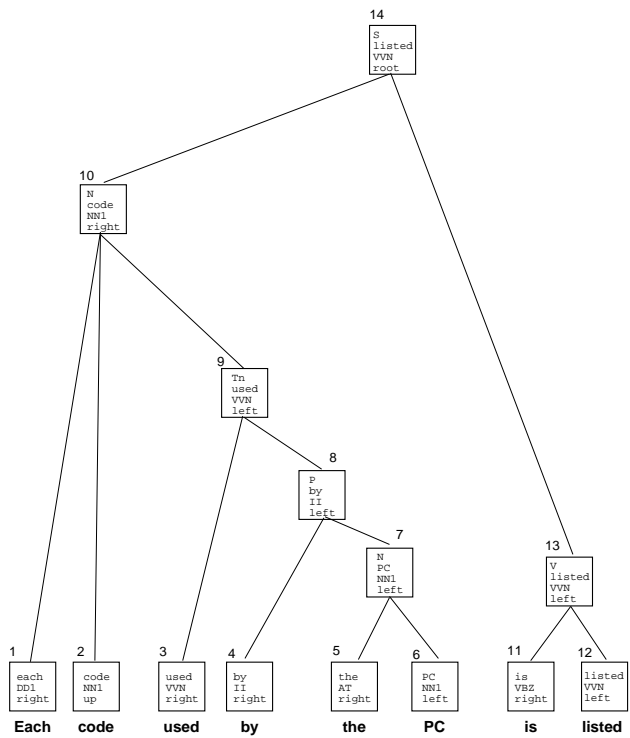,width=3.25in}}
\caption{Treebank analysis encoded using feature values.\label{fullparse}}
\end{figure}
\end{centering}

The SPATTER representation of the sentence
\begin{verbatim}
(S (N Each_DD1 code_NN1
      (Tn used_VVN
          (P by_II (N the_AT PC_NN1))))
   (V is_VBZ listed_VVN))
\end{verbatim}
is shown in Figure~\ref{fullparse}.  The nodes are constructed
bottom-up from left-to-right, with the constraint that no constituent
node is constructed until all of its children have been constructed.
The order in which the nodes of the example sentence are constructed
is indicated in the figure.

\subsection{Training SPATTER's models}

SPATTER consists of three main decision-tree models: a part-of-speech
tagging model, a node-extension model, and a node-labeling model.

Each of these decision-tree models are grown using the following
questions, where $X$ is one of word, tag, label, or extension, and $Y$
is either left and right:
\begin{itemize}
\item What is the $X$ at the current node?
\item What is the $X$ at the node to the $Y$?
\item What is the $X$ at the node two nodes to the $Y$?
\item What is the $X$ at the current node's first child from the $Y$?
\item What is the $X$ at the current node's second child from the $Y$?
\end{itemize}
For each of the nodes listed above, the decision tree could also ask
about the number of children and span of the node.  For the tagging
model, the values of the previous two words and their tags are also
asked, since they might differ from the head words of the previous two
constituents.

The training algorithm proceeds as follows.  The training corpus is
divided into two sets, approximately 90\% for tree growing and 10\%
for tree smoothing.  For each parsed sentence in the tree growing
corpus, the correct state sequence is traversed.  Each state
transition from $s_i$ to $s_{i+1}$ is an event; the history is made up
of the answers to all of the questions at state $s_i$ and the future
is the value of the action taken from state $s_i$ to state $s_{i+1}.$
Each event is used as a training example for the decision-tree growing
process for the appropriate feature's tree (e.g. each tagging event is
used for growing the tagging tree, etc.).  After the decision trees
are grown, they are smoothed using the tree smoothing corpus using a
variation of the deleted interpolation algorithm described in
\cite{mythesis}.

\subsection{Parsing with SPATTER}

The parsing procedure is a search for the highest probability parse
tree.  The probability of a parse is just the product of the
probability of each of the actions made in constructing the parse,
according to the decision-tree models.

Because of the size of the search space, (roughly $O(|T|^n|N|^n),$
where $|T|$ is the number of part-of-speech tags, $n$ is the number of
words in the sentence, and $|N|$ is the number of non-terminal
labels), it is not possible to compute the probability of every parse.
However, the specific search algorithm used is not very important, so
long as there are no search errors.  A search error occurs when the
the highest probability parse found by the parser is not the highest
probability parse in the space of all parses.

SPATTER's search procedure uses a two phase approach to identify the
highest probability parse of a sentence.  First, the parser uses a
stack decoding algorithm to quickly find a complete parse for the
sentence.  Once the stack decoder has found a complete parse of
reasonable probability ($>10^{-5}$), it switches to a breadth-first
mode to pursue all of the partial parses which have not been explored
by the stack decoder.  In this second mode, it can safely discard any
partial parse which has a probability lower than the probability of
the highest probability completed parse.  Using these two search
modes, SPATTER guarantees that it will find the highest probability
parse.  The only limitation of this search technique is that, for
sentences which are modeled poorly, the search might exhaust the
available memory before completing both phases.  However, these search
errors conveniently occur on sentences which SPATTER is likely to get
wrong anyway, so there isn't much performance lossed due to the search
errors.  Experimentally, the search algorithm guarantees the highest
probability parse is found for over 96\% of the sentences parsed.

\section{Experiment Results}

In the absence of an NL system, SPATTER can be evaluated by comparing
its top-ranking parse with the treebank analysis for each test
sentence.  The parser was applied to two different domains, IBM
Computer Manuals and the Wall Street Journal.

\subsection{IBM Computer Manuals}

The first experiment uses the IBM Computer Manuals domain, which
consists of sentences extracted from IBM computer manuals.  The
training and test sentences were annotated by the University of
Lancaster.  The Lancaster treebank uses 195 part-of-speech tags and 19
non-terminal labels.  This treebank is described in great detail in
\cite{blackbook}.

The main reason for applying SPATTER to this domain is that IBM had
spent the previous ten years developing a rule-based,
unification-style probabilistic context-free grammar for parsing this
domain.  The purpose of the experiment was to estimate SPATTER's
ability to learn the syntax for this domain directly from a treebank,
instead of depending on the interpretive expertise of a grammarian.

The parser was trained on the first 30,800 sentences from the
Lancaster treebank.  The test set included 1,473 new sentences, whose
lengths range from 3 to 30 words, with a mean length of 13.7 words.
These sentences are the same test sentences used in the experiments
reported for IBM's parser in \cite{blackbook}.  In \cite{blackbook},
IBM's parser was evaluated using the 0-crossing-brackets measure,
which represents the percentage of sentences for which none of the
constituents in the parser's parse violates the constituent boundaries
of any constituent in the correct parse.  After over ten years of
grammar development, the IBM parser achieved a 0-crossing-brackets
score of 69\%.  On this same test set, SPATTER scored 76\%.

\subsection{Wall Street Journal}

The experiment is intended to illustrate SPATTER's ability to
accurately parse a highly-ambiguous, large-vocabulary domain.  These
experiments use the Wall Street Journal domain, as annotated in the
Penn Treebank, version 2.  The Penn Treebank uses 46 part-of-speech
tags and 27 non-terminal labels.\footnote{This treebank also contains
coreference information, predicate-argument relations, and trace
information indicating movement; however, none of this additional
information was used in these parsing experiments.}

The WSJ portion of the Penn Treebank is divided into 25 sections,
numbered 00 - 24.  In these experiments, SPATTER was trained on
sections 02 - 21, which contains approximately 40,000 sentences.  The
test results reported here are from section 00, which contains 1920
sentences.\footnote{For an independent research project on
coreference, sections 00 and 01 have been annotated with detailed
coreference information.  A portion of these sections is being used as
a development test set.  Training SPATTER on them would improve
parsing accuracy significantly and skew these experiments in favor of
parsing-based approaches to coreference.  Thus, these two sections
have been excluded from the training set and reserved as test
sentences.}  Sections 01, 22, 23, and 24 will be used as test data in
future experiments.

The Penn Treebank is already tokenized and sentence detected by human
annotators, and thus the test results reported here reflect this.
SPATTER parses {\em word} sequences, not tag sequences.  Furthermore,
SPATTER does {\em not} simply pre-tag the sentences and use only the
best tag sequence in parsing.  Instead, it uses a probabilistic model
to assign tags to the words, and considers all possible tag sequences
according to the probability they are assigned by the model.  No
information about the legal tags for a word are extracted from the
test corpus.  In fact, no information other than the words is used
from the test corpus.

For the sake of efficiency, only the sentences of 40 words or fewer
are included in these experiments.\footnote{SPATTER returns a complete
parse for all sentences of fewer then 50 words in the test set, but
the sentences of 41 - 50 words required much more computation than the
shorter sentences, and so they have been excluded.}  For this test
set, SPATTER takes on average 12 seconds per sentence on an SGI R4400
with 160 megabytes of RAM.

To evaluate SPATTER's performance on this domain, I am using the
PARSEVAL measures, as defined in \cite{CBpaper}:
\begin{description}
\item[Precision] $$\frac{\mbox{no. of correct
constituents in SPATTER parse}}{\mbox{no. of constituents in SPATTER
parse}}$$
\item[Recall] $$\frac{\mbox{no. of correct
constituents in SPATTER parse}}{\mbox{no. of constituents in treebank
parse}}$$
\item[Crossing Brackets] no. of constituents which violate constituent
boundaries with a constituent in the treebank parse.
\end{description}

The precision and recall measures do not consider constituent labels
in their evaluation of a parse, since the treebank label set will not
necessarily coincide with the labels used by a given grammar.  Since
SPATTER uses the same syntactic label set as the Penn Treebank, it
makes sense to report labelled precision and labelled recall.  These
measures are computed by considering a constituent to be correct if
and only if it's label matches the label in the treebank.

Table~\ref{WSJRESULTStable} shows the results of SPATTER evaluated
against the Penn Treebank on the Wall Street Journal section 00.


\begin{table}[tbhp]
\begin{center}
\begin{tabular}{|l|c|c|c|c|}
\hline
\hline
Sent. Length Range &			4-40   & 4-25 & 10-20\\
\hline
\hline
Comparisons  &				1759   & 1114 & 653\\
Avg. Sent. Length &			22.3   & 16.8 & 15.6\\
Treebank Constituents &			17.58  & 13.21 & 12.10\\
Parse Constituents  &			17.48  & 13.13 & 12.03\\
\hline
Tagging Accuracy  &			96.5\% & 96.6\% & 96.5\%\\
\hline
Crossings Per Sentence &		1.33   & 0.63 & 0.49\\
Sent. with 0 Crossings &		55.4\% & 69.8\% & 73.8\%\\
Sent. with 1 Crossing	 &		69.2\% & 83.8\% & 86.8\%\\
Sent. with 2 Crossings	 &		80.2\% & 92.1\% & 95.1\%\\
\hline
Precision 		 &		86.3\% & 89.8\% & 90.8\%\\
Recall  &				85.8\% & 89.3\% & 90.3\%\\
Labelled Precision 	 &		84.5\% & 88.1\% & 89.0\%\\
Labelled Recall  &			84.0\% & 87.6\% & 88.5\%\\
\hline
\hline
\end{tabular}
\end{center}
\caption{Results from the WSJ Penn Treebank
experiments.\label{WSJRESULTStable}}
\end{table}

\begin{centering}
\begin{figure}[tbhp]
\centerline{\psfig{figure=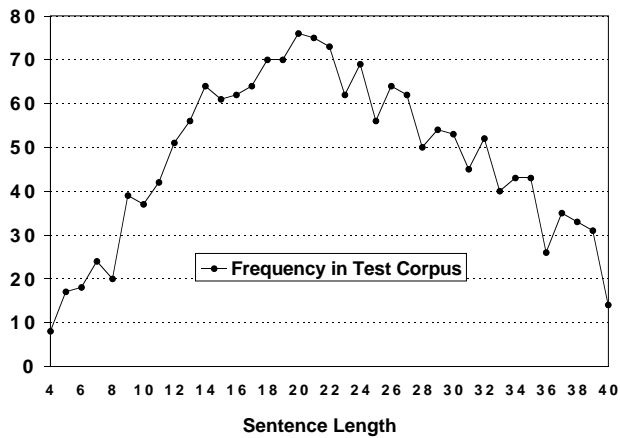,width=3.25in}}
\caption{Frequency in the test corpus as a function of sentence length for
Wall Street Journal experiments.\label{WSJFrequency}}
\end{figure}
\end{centering}

\begin{centering}
\begin{figure}[tbhp]
\centerline{\psfig{figure=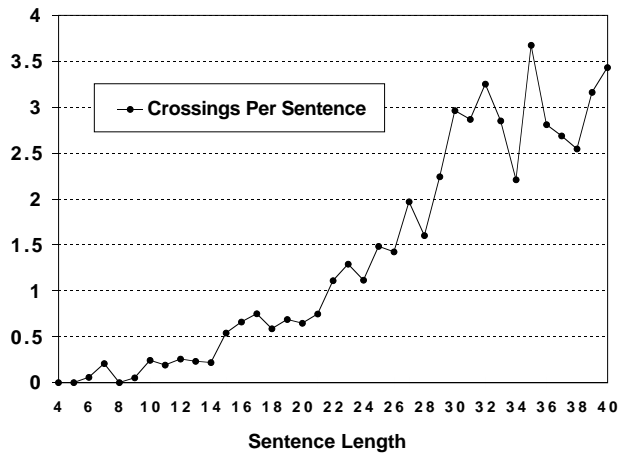,width=3.25in}}
\caption{Number of crossings per sentence as a function of sentence
length for Wall Street Journal experiments.\label{WSJCrossings}}
\end{figure}
\end{centering}

\begin{centering}
\begin{figure}[tbhp]
\centerline{\psfig{figure=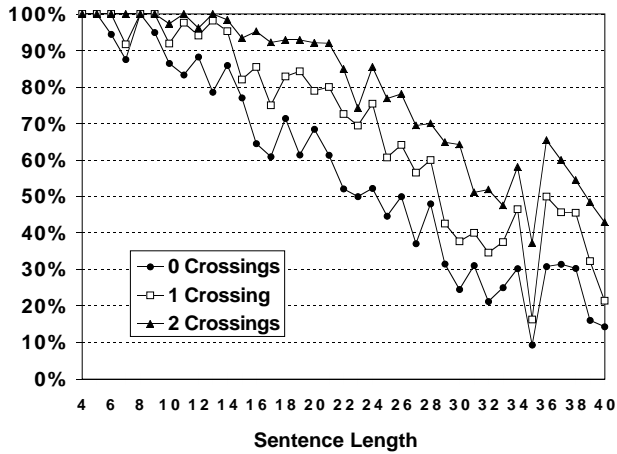,width=3.25in}}
\caption{Percentage of sentence with 0, 1, and 2 crossings as a
function of sentence length for Wall Street Journal
experiments.\label{WSJ0Crossing}}
\end{figure}
\end{centering}

\begin{centering}
\begin{figure}[tbhp]
\centerline{\psfig{figure=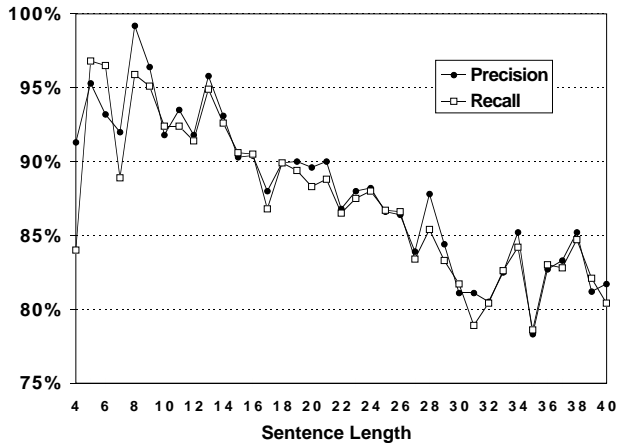,width=3.25in}}
\caption{Precision and recall as a function of sentence length for
Wall Street Journal experiments.\label{WSJPrecisionRecall}}
\end{figure}
\end{centering}

Figures~\ref{WSJCrossings}, \ref{WSJ0Crossing}, and
\ref{WSJPrecisionRecall} illustrate the performance of SPATTER as a
function of sentence length.  SPATTER's performance degrades slowly
for sentences up to around 28 words, and performs more poorly and more
erratically as sentences get longer.  Figure~\ref{WSJFrequency}
indicates the frequency of each sentence length in the test corpus.

\section{Conclusion}

Regardless of what techniques are used for parsing disambiguation, one
thing is clear: if a particular piece of information is necessary for
solving a disambiguation problem, it must be made available to the
disambiguation mechanism.  The words in the sentence are clearly
necessary to make parsing decisions, and in some cases long-distance
structural information is also needed.  Statistical models for parsing
need to consider many more features of a sentence than can be managed
by {$n$}-gram modeling techniques and many more examples than a human
can keep track of.  The SPATTER parser illustrates how large amounts
of contextual information can be incorporated into a statistical model
for parsing by applying decision-tree learning algorithms to a large
annotated corpus.

\end{document}